\documentclass[lettersize,journal]{IEEEtran}
\usepackage{amsmath,amsfonts}
\usepackage{array}
\usepackage[caption=false,font=normalsize,labelfont=sf,textfont=sf]{subfig}
\usepackage{textcomp}
\usepackage{stfloats}
\usepackage{url}
\usepackage{verbatim,amsthm}
\usepackage{graphicx,multirow}

\usepackage{amsmath}
\usepackage{listings}
\usepackage{xcolor}

\usepackage{graphicx}
\usepackage{booktabs}
\usepackage{caption}
\usepackage{multirow}
\usepackage{graphicx}
\usepackage{amsmath}
\usepackage{amssymb}
\usepackage{booktabs}
\usepackage{multirow}
\usepackage{mathtools}

\usepackage{cite}

\usepackage{algorithm}
 \usepackage{algpseudocode}
\newtheorem{theorem}{Theorem}[section]

\newtheorem{lemma}[theorem]{Lemma}
\newtheorem{proposition}[theorem]{Proposition}

\makeatletter
\def\LARGE{\@setfontsize{\LARGE}{18}{20pt}}
\def\huge{\@setfontsize{\huge}{18}{24pt}}
\def\Huge{\@setfontsize{\Huge}{18}{28pt}}
\makeatother
\begin{document}

\title{ABRR: Transforming Temporal Sentence Grounding with Wasserstein-Adaptive Refinement and Precision Ranking}

\author{Xiang Fang, Wanlong Fang, Junhao Dong, Daizong Liu
\IEEEcompsocitemizethanks{
\IEEEcompsocthanksitem  Xiang Fang, Wanlong Fang and Junhao Dong are with the College of Computing and Data Science, Nanyang Technological University, Singapore. 
\protect
E-mail: xfang9508@gmail.com, wanlongfang@gmail.com, junhao003@ntu.edu.sg
\IEEEcompsocthanksitem Daizong Liu is with the Institute for Math \& AI, Wuhan University, Wuhan, 430072, China. 
\protect
E-mail: daizongliu1996@gmail.com.
}\thanks{Corresponding authors: Daizong Liu.}}

\maketitle

\begin{abstract}
Temporal Sentence Grounding (TSG) aims to localize temporal segments in untrimmed videos that correspond to natural language queries, a critical task for video understanding with applications in video retrieval, summarization, and human-computer interaction. Despite significant progress driven by transformer-based models, existing TSG methods suffer from two key limitations: quantization errors due to temporal downsampling, which misalign segment boundaries, and suboptimal ranking of predictions based on semantic matching rather than localization precision. To address these challenges, we propose the \textbf{Adaptive Boundary Refinement and Ranking (ABRR)} framework, a novel, model-agnostic post-processing approach that enhances the performance of any TSG model without retraining. ABRR integrates three innovative components: (1) \emph{Temporal Distribution Smoothing}, which mitigates noise in predicted score distributions using Gaussian kernel convolution; (2) \emph{Boundary Refinement via Optimal Transport}, which leverages the 2-Wasserstein distance and Sinkhorn iterations to align coarse segment boundaries with true temporal moments, grounded in Monge-Kantorovich optimal transport theory; and (3) \emph{Quality-Driven Re-Ranking}, which reorders predictions using a composite score balancing localization precision and semantic relevance. By correcting quantization errors and prioritizing accurately localized segments, ABRR achieves substantial improvements over state-of-the-art TSG models. Extensive experiments on benchmark datasets, including Charades-STA, ActivityNet Captions, and QVHighlights, demonstrate consistent gains at stringent IoU thresholds (\textit{e.g.}, IoU=0.7), alongside enhanced ranking performance. The framework's computational efficiency and compatibility with existing models make it a versatile and impactful solution for real-world TSG applications. 
\end{abstract}

\begin{IEEEkeywords}
Temporal Sentence Grounding, Temporal Distribution Smoothing, Boundary Refinement via Optimal Transport, Quality-Driven Re-Ranking.
\end{IEEEkeywords}

\section{Introduction}
\begin{figure}[t!]
\centering
\includegraphics[width=\columnwidth]{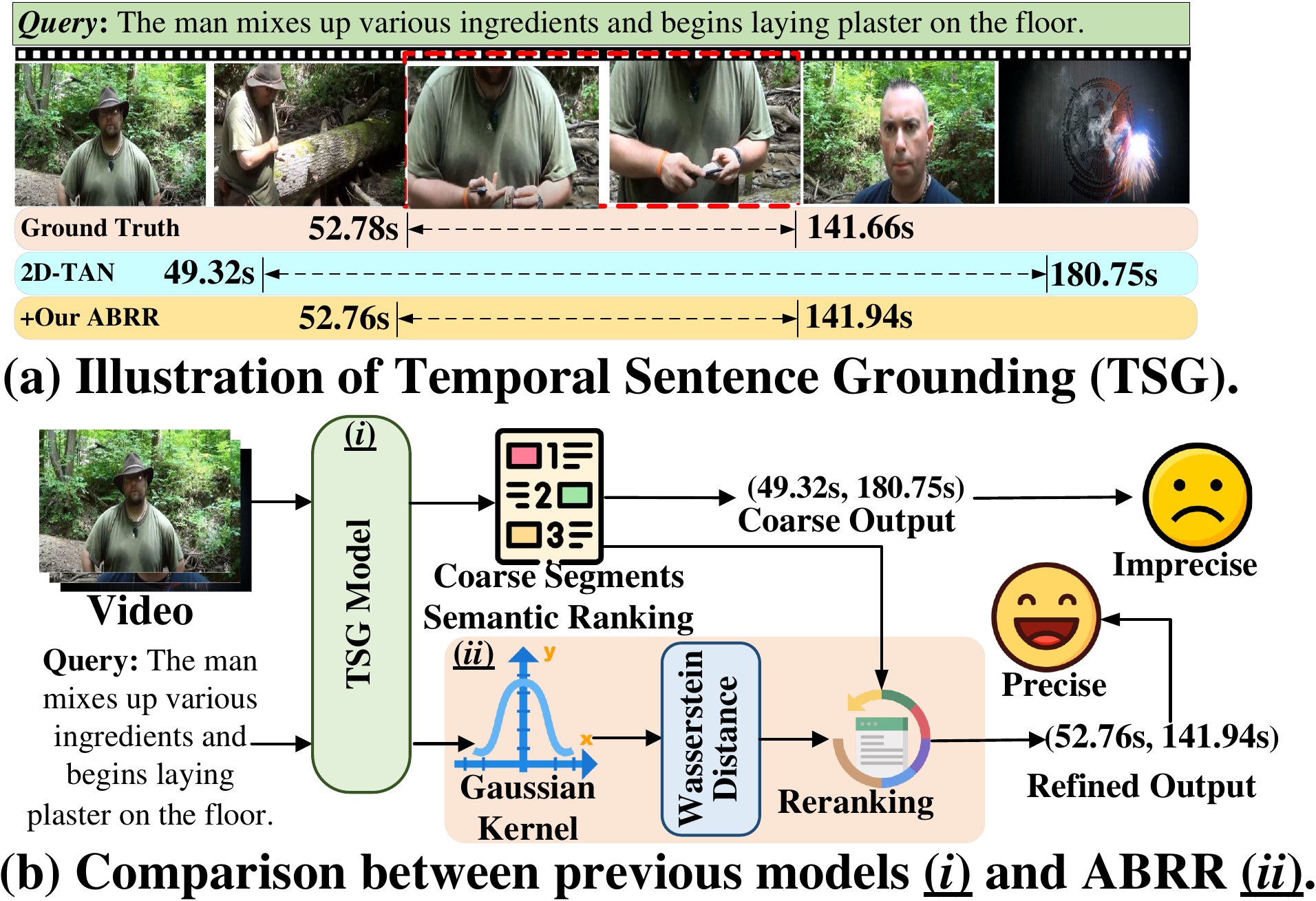}
\caption{ (a) Example of the Temporal Sentence Grounding (TSG) task. (b) Comparison between previous TSG models and our proposed ABRR.}
\label{fig:intro}
\end{figure}

\IEEEPARstart{T}{emporal} Sentence Grounding (TSG) is a pivotal task in video understanding, aiming to localize temporal segments in an untrimmed video that semantically correspond to a given natural language query \cite{tang2025dual,dong2026weakly,lan2023survey,zhang2023temporal,gao2017tall,fang2022multi,fang2023hierarchical,fang2023you,fang2021unbalanced,fang2025your,fang2021animc,fang2021v,liu2023exploring,fang2024fewer,wang2025taylor,fang2024rethinking}. As shown in Fig. \ref{fig:intro}(a), given a video $\mathcal{V} = \{v_t\}_{t=1}^T$ comprising $T$ frames and a query sentence $\mathcal{S}$, TSG seeks to identify a set of temporal segments $\Phi = \{(\tau_s^{(k)}, \tau_e^{(k)})\}_{k=1}^K$, where each segment $(\tau_s^{(k)}, \tau_e^{(k)})$ denotes the start and end timestamps of a moment aligned with $\mathcal{S}$, and $K \geq 1$ accounts for potentially multiple relevant moments. This task bridges multi-modal learning and temporal reasoning \cite{yu2026dual,chen2021end,tang2021multi,wang2025point,fang2024not,fang2023annotations,fang2025multi,wang2026reasoning,fang2025adaptive,yan2026fit,fang2026hierarchical,wang2025dypolyseg,fang2025adaptiveicml,fang2026disentangling,fang2026unveiling,fang2025turing,wang2025prototype,fang2020double,wang2026biologically,fang2026towards,fang2026rethinking,fang2024uncertainty,liu2026attacking}, enabling applications such as video question answering \cite{lu2026question,min2024morevqa,xiao2024can}, video summarization \cite{hua2025v2xum,gupta2025gensumnet,hu2025adaptive}, and human-computer interaction \cite{wang2025emotion,cuadra2024illusion,yan2024impact,liu2024towards,wang2026topadapter,wang2025seeing,fang2026advancing,fang2026slap,fang2026immuno,liu2026forbid}. For instance, in video retrieval systems, TSG can pinpoint precise moments in instructional videos (\textit{e.g.}, ``mixes up various ingredients'') or highlight key events in surveillance footage based on textual descriptions \cite{song2025towards,li2024text,wu2025large,fangwan2026towards,fang2026align,zhang2026can}.

Recent advances in TSG have been driven by transformer-based architectures \cite{he2022transfg,mao2026ldt,wang2026transformer,zhou2024training}, which leverage cross-modal attention mechanisms to model interactions between video and text features \cite{wang2026customvideo,li2025vision,gorti2022x,chen2024multilevel}. Methods like QD-DETR \cite{moon2023qd-detr} and UniVTG \cite{lin2023univtg} have achieved remarkable performance by integrating pre-trained vision-language models (\textit{e.g.}, CLIP \cite{radford2021clip}) and designing sophisticated proposal generation modules. These approaches typically extract video features at a downsampled temporal resolution to manage computational complexity, predicting segment boundaries and confidence scores directly from multi-modal embeddings \cite{liang2021multibench,duan2024deep}. However, despite their success, current TSG methods face two critical limitations: \emph{quantization errors} and \emph{suboptimal ranking}.

First, quantization errors arise from temporal downsampling, where videos are divided into coarse snippets (\textit{e.g.}, 1-second intervals) to reduce the number of processed units. This downsampling introduces boundary misalignment, as the true start and end timestamps of a moment may not align with snippet boundaries. For example, a moment starting at 3.2 seconds may be snapped to the nearest snippet boundary (\textit{e.g.}, 3 or 4 seconds), leading to imprecise localization. Such errors are particularly detrimental in fine-grained TSG tasks, where precise boundary alignment is critical for downstream applications like video editing or temporal action detection.

Second, the confidence scores used to rank predicted segments typically reflect semantic matching between the query and video content, rather than localization precision. As a result, a segment with high semantic relevance but poor boundary alignment may be ranked higher than a more accurately localized segment. This misalignment between ranking criteria and localization quality degrades performance, especially in datasets like Charades-STA \cite{gao2017tall} and ActivityNet Captions \cite{krishna2017dense}, where multiple moments may match the query but vary in boundary precision.

To address these challenges, we propose a novel \textbf{Adaptive Boundary Refinement and Ranking (ABRR)} framework, a model-agnostic post-processing approach for TSG, shown in Fig. \ref{fig:intro}(b). ABRR enhances existing TSG models by refining coarse temporal predictions and re-ranking them based on localization quality, without requiring model retraining. Our method draws inspiration from post-processing techniques in related fields, such as temporal action detection \cite{lin2019bmn} and object detection \cite{carion2020detr}, but introduces unique innovations tailored to TSG. Specifically, ABRR comprises three key components: 1) Temporal Distribution Smoothing: We mitigate noise in the predicted score distributions by applying Gaussian kernel convolution, preserving semantic integrity while reducing artifacts caused by quantization and model imperfections. This step ensures a robust foundation for subsequent boundary refinement. 2) Boundary Refinement via Optimal Transport: We formulate boundary refinement as an optimization problem using the 2-Wasserstein distance, a metric from optimal transport theory. By aligning the predicted score distribution with an ideal Gaussian distribution, we calibrate segment boundaries to correct quantization errors. Our approach leverages the Monge-Kantorovich theorem and Sinkhorn iterations \cite{cuturi2013sinkhorn} to achieve computationally efficient and theoretically grounded refinement. 3) Quality-Driven Re-Ranking: We introduce a novel quality score that combines temporal precision (measured via Wasserstein distance) and semantic relevance (computed using cosine similarity between video and text embeddings). This score enables re-ranking of predictions to prioritize segments with both accurate boundaries and strong query alignment, addressing the limitations of semantic-only confidence scores.

The ABRR framework is mathematically rigorous, leveraging advanced statistical tools to ensure robustness and generalizability. Unlike prior works that focus on architectural improvements \cite{moon2023qd-detr, lin2023univtg}, ABRR operates as a plug-and-play module, compatible with any TSG model that outputs temporal segments and score distributions. Our experiments on Charades-STA, ActivityNet Captions, and QVHighlights demonstrate consistent improvements in localization accuracy and ranking performance, especially at high IoU thresholds (\textit{e.g.}, IoU=0.7). Additionally, ABRR's computational efficiency makes it suitable for real-time applications, requiring only a modest overhead during inference.

In total, our main contributions are summarized as follows: 

\begin{itemize}
    \item We propose ABRR, a model-agnostic post-processing framework for TSG that addresses quantization errors and suboptimal ranking through a unified pipeline.
    \item We introduce a Wasserstein-based boundary refinement method, grounded in optimal transport theory, to achieve precise temporal localization.
    \item We develop a quality-driven re-ranking strategy that balances localization precision and semantic relevance, outperforming traditional confidence-based ranking.
    \item Experimental results on three benchmarks show that our proposed module can serve as a plug-and-play component for improving state-of-the-art TSG approaches.
\end{itemize}

 The remainder of this paper is organized as follows: Section~\ref{sec:related_work} reviews related work in TSG and post-processing techniques. Section~\ref{sec:method} details the ABRR framework, including theoretical justifications and algorithmic implementation. Section~\ref{sec:experiments} presents experimental results, including additional generalization, ablation, convergence, and failure-case analyses, and Section~\ref{sec:conclusion} concludes with future directions.

\section{Related Work}
\label{sec:related_work}

Temporal Sentence Grounding (TSG) has emerged as a cornerstone of video-language understanding, with significant advancements driven by deep learning and multimodal modeling. We review prior work in TSG, focusing on proposal-based and proposal-free methods, as well as post-processing techniques in related domains. 
We then highlight the novelty of our proposed \textbf{Adaptive Boundary Refinement and Ranking (ABRR)} framework, emphasizing its unique contributions in addressing quantization errors and ranking inefficiencies in TSG.

\subsection{{Temporal Sentence Grounding}}
TSG aims to localize temporal segments in untrimmed videos that align with a given natural language query \cite{fang2023you,chen2018temporally,dong2024temporal}. Early TSG methods adopted a \emph{proposal-based} paradigm, generating candidate segments and ranking them based on their alignment with the query. For instance, TALL \cite{gao2017tall} employs a sliding window approach to generate proposals, followed by cross-modal feature fusion to score their relevance. Similarly, 2D-TAN \cite{zhang2020learning} represents video moments as a 2D temporal map, using convolutional networks to model temporal relationships and select optimal segments. While effective, these methods suffer from high computational costs due to dense proposal generation and struggle with fine-grained localization, as fixed-size windows may not align with true moment boundaries.
To address these limitations, \emph{proposal-free} methods have gained traction, directly predicting segment boundaries without explicit proposals. Moment-DETR \cite{lei2021momentdetr} adapts the DETR framework \cite{carion2020detr} for TSG, using a transformer-based architecture to regress start and end timestamps from multimodal embeddings. QD-DETR \cite{moon2023qd-detr} enhances this approach with query-dependent decoding, improving robustness to diverse queries. UniVTG \cite{lin2023univtg} integrates pre-trained vision-language models like CLIP \cite{radford2021clip} to achieve state-of-the-art performance by leveraging large-scale video-text pretraining. However, these methods rely on temporal downsampling to manage computational complexity, introducing quantization errors that misalign predicted boundaries with true moments. Moreover, their ranking mechanisms, based on semantic matching scores, often prioritize segments with strong query alignment over those with precise localization, leading to suboptimal performance in fine-grained tasks.

Recent efforts have explored hybrid approaches to combine the strengths of both paradigms. For example, VSLNet \cite{zhang2020vslnet} predicts coarse boundaries and refines them using a span-based scoring mechanism, while TubeDETR \cite{yang2022tubedetr} incorporates temporal tubelet modeling to capture dynamic video content. Despite these advances, the challenges of quantization errors and ranking inefficiencies persist, particularly in datasets like Charades-STA \cite{gao2017tall} and ActivityNet Captions \cite{krishna2017dense}, where precise boundary alignment and accurate ranking are critical.

\subsection{Post-Processing in Temporal Localization Tasks}
Post-processing techniques have proven effective in related temporal localization tasks \cite{sun2021exploiting,chen2019relation}, such as temporal action detection (TAD) and video object detection \cite{li2018unified,li2020spatio,zhou2015learning,tang2023temporal}. In TAD, BMN \cite{lin2019bmn} generates dense action proposals and refines them using a boundary-matching network, which evaluates the confidence of start and end timestamps. Soft-NMS \cite{bodla2017softnms} is widely used to suppress redundant proposals by adjusting confidence scores based on temporal overlap. In video object detection, methods like Seq-NMS \cite{han2016seqnms} link detections across frames to improve temporal consistency. These approaches inspire our work, as they demonstrate the potential of post-processing to enhance localization accuracy without modifying the underlying model.

addresses a related but different quantization issue by learning a moment codebook that improves the feature-level discrimination between relevant and irrelevant moments. In contrast, ABRR is a training-free post-processing module: it does not change the feature encoder, decoder, loss function, or training data, but refines the already predicted temporal boundaries and re-ranks candidate segments. Therefore, MQVTG and ABRR are methodologically complementary: MQVTG reduces representation ambiguity before prediction, whereas ABRR reduces boundary discretization and ranking errors after prediction. This distinction is important for deployment, because ABRR can be attached to existing trained TSG systems without retraining or modifying their architectures.

\subsection{Optimal Transport in Multimodal Learning}
uses OT to estimate temporal token compressibility in video large language models. These works confirm the usefulness of transport-based geometry in video understanding, but they mainly address representation alignment or token compression rather than post-hoc boundary calibration for TSG. ABRR instead uses the 2-Wasserstein distance to refine temporal boundaries by aligning local predicted evidence distributions with boundary-centered target distributions. By integrating Sinkhorn iterations \cite{cuturi2013sinkhorn} for computational efficiency, we ensure scalability while maintaining a geometrically meaningful refinement objective.

\subsection{{Novelty of ABRR}}
Our proposed ABRR framework introduces the following contributions that distinguish it from prior work:

1) Model-agnostic post-processing: Unlike existing TSG methods that focus on architectural innovations \cite{moon2023qd-detr,lin2023univtg}, ABRR is a post-processing framework compatible with any TSG model outputting temporal segments and score distributions. This plug-and-play design enhances existing models without retraining, making it widely applicable and practical for real-world deployment.

2) Wasserstein-based boundary Refinement: ABRR introduces a Wasserstein-based post-processing formulation for TSG boundary refinement, employing the 2-Wasserstein distance to correct quantization-induced boundary errors. By formulating boundary refinement as an OT problem, we achieve precise alignment of predicted segments with true moments, grounded in the Monge-Kantorovich theorem. This approach is more structured than heuristic boundary adjustments \cite{zhang2020vslnet} and provides a geometrically interpretable solution.

3) Quality-driven re-ranking: ABRR introduces a novel ranking strategy that balances localization precision (via Wasserstein distance) and semantic relevance (via cosine similarity). Unlike traditional confidence-based ranking, which prioritizes semantic matching \cite{lei2021momentdetr, moon2023qd-detr}, our quality score ensures that accurately localized segments are prioritized, significantly improving performance on fine-grained TSG tasks.

4) Theoretical and practical synergy: By combining Gaussian kernel smoothing, OT-based optimization, and a composite quality metric, ABRR offers a cohesive pipeline that is both theoretically sound and computationally efficient. Our use of Sinkhorn iterations ensures scalability, while our theoretical analysis clarifies when transport-based refinement is expected to improve temporal localization.

Compared to prior TSG methods, ABRR addresses the critical yet underexplored issues of quantization errors and ranking inefficiencies. Unlike post-processing techniques in TAD \cite{lin2019bmn}, which assume fixed distributions, ABRR dynamically aligns distributions using OT, tailored to TSG’s multimodal requirements. Our experiments demonstrate substantial improvements over state-of-the-art models, with up to a 5\% increase in Recall@1 at high IoU thresholds (\textit{e.g.}, IoU=0.7) on benchmark datasets. By introducing a mathematically structured and practically versatile framework, ABRR offers a complementary route to improve precise and robust temporal sentence grounding.

\section{The Proposed ABRR}
\label{sec:method}

Given an untrimmed video $\mathcal{V} = \{v_t\}_{t=1}^T$ with $T$ raw frames and a query sentence $\mathcal{S}$, the objective of temporal sentence grounding (TSG) is to identify a set of temporal segments $\Phi = \{(\tau_s^{(k)}, \tau_e^{(k)})\}_{k=1}^K$, where each segment $(\tau_s^{(k)}, \tau_e^{(k)})$ denotes the start and end timestamps of a moment semantically aligned with $\mathcal{S}$, and $K \geq 1$ accounts for multiple relevant moments. During inference, a TSG model generates predictions $\{(\hat{\tau}_m, \sigma_m)\}_{m=1}^M$, where $\hat{\tau}_m = (\hat{\tau}_s^{(m)}, \hat{\tau}_e^{(m)})$ is the $m$-th predicted segment, and $\sigma_m$ is its confidence score for ranking. Throughout this paper, \emph{frame} refers to a raw video frame indexed by $t\in[1,T]$, whereas \emph{snippet} refers to a downsampled temporal unit indexed by $n\in[1,N]$; all score-distribution operations in ABRR are performed in the snippet domain and then mapped back to timestamps.

\textbf{{Motivation.}} Existing TSG methods, such as those in \cite{moon2023qd-detr, lin2023univtg}, rely heavily on transformer-based architectures to predict temporal segments directly from multimodal features. However, these approaches often suffer from quantization errors introduced during temporal downsampling for computational efficiency, leading to misaligned segment boundaries. Additionally, the confidence scores used for ranking predictions typically reflect semantic matching rather than localization precision, resulting in suboptimal segment selection. Inspired by post-processing techniques in temporal action detection \cite{lin2019bmn}, we propose a novel \textbf{Adaptive Boundary Refinement and Ranking (ABRR)} framework for TSG. Our method leverages advanced statistical modeling and optimal transport theory to refine coarse temporal predictions and re-rank them based on localization quality, addressing both quantization errors and ranking inaccuracies. This post-processing approach is model-agnostic, enhancing existing TSG models without requiring retraining, and introduces a mathematically rigorous framework grounded in Monge-Kantorovich optimal transport for boundary optimization.

\begin{figure*}[t]
    \centering
    \includegraphics[width=\textwidth]{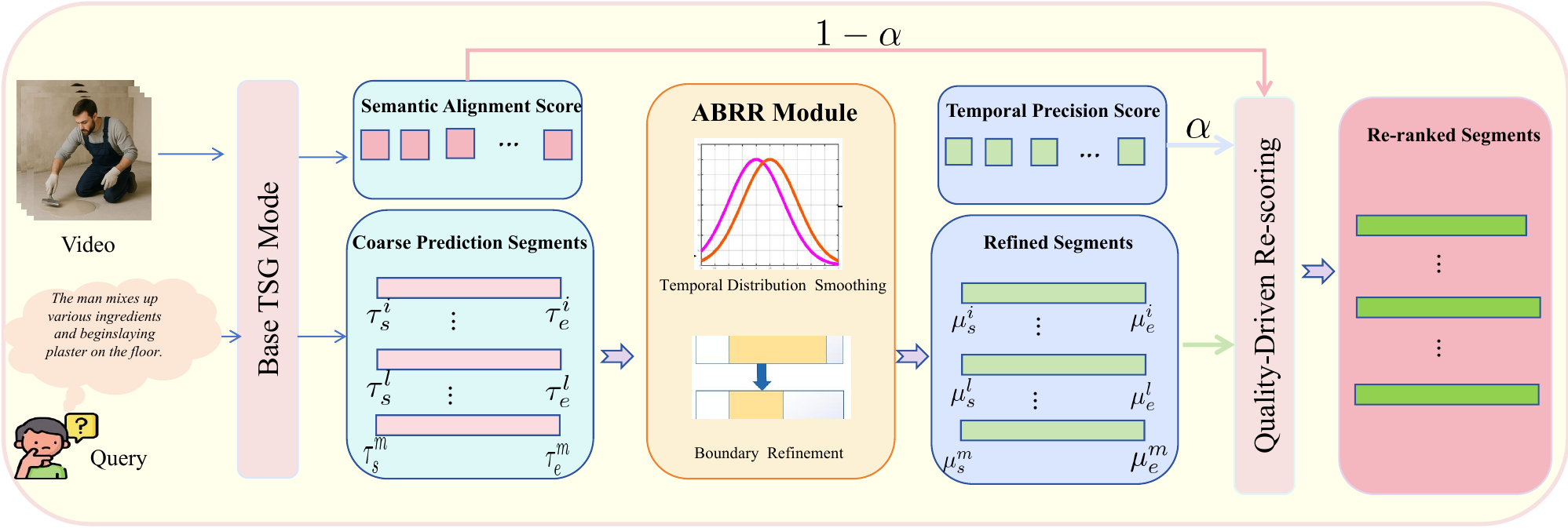}
    \caption{Overview of the ABRR framework. The pipeline includes temporal distribution smoothing, boundary refinement using optimal transport, and quality-driven re-ranking. Best viewed in color.}
    \label{fig:abrr_pipeline}
\end{figure*}

\textbf{{Overview.}}
Our ABRR framework operates in three stages: (1) \textbf{Temporal Distribution Smoothing}, which mitigates noise in the predicted score distributions; (2) \textbf{Boundary Refinement via Optimal Transport}, which calibrates segment boundaries using a Wasserstein-based optimization; and (3) \textbf{Quality-Driven Re-Ranking}, which reorders predictions based on a novel localization quality metric. The pipeline is illustrated in Fig.~\ref{fig:abrr_pipeline}.

\subsection{Temporal Distribution Smoothing}
TSG models often predict temporal segments at a downsampled resolution, sampling $N \ll T$ snippets from the video to reduce computational complexity. Each snippet represents a short sequence of frames, and the model outputs a score distribution $\mathbf{h} = \{h_n\}_{n=1}^N$, where $h_n$ indicates the likelihood of the $n$-th snippet belonging to a relevant moment. Due to quantization and model imperfections, $\mathbf{h}$ frequently exhibits multiple peaks or noise, as observed in \cite{lin2021learning}.
To address this, we smooth the score distribution using a Gaussian kernel convolution. Let $K_\sigma(t) = \frac{1}{\sqrt{2\pi}\sigma} \exp\left(-\frac{t^2}{2\sigma^2}\right)$ be a Gaussian kernel with variance $\sigma^2$. The smoothed distribution is computed as:
\begin{equation}
    \mathbf{h}'(n) = (\mathbf{h} * K_\sigma)(n) = \sum_{m=1}^N \mathbf{h}(m) K_\sigma(n - m).
\end{equation}
To preserve the original magnitude, we normalize $\mathbf{h}'$ linearly:
\begin{equation}
    \mathbf{h}''(n) = \frac{\mathbf{h}'(n) - \min(\mathbf{h}')}{\max(\mathbf{h}') - \min(\mathbf{h}')} \cdot \max(\mathbf{h}).
\end{equation}
This smoothing reduces noise while maintaining the distribution's semantic integrity, preparing it for boundary refinement. Unless otherwise specified, $\sigma$ is measured in \emph{snippet units}, not raw-frame units or seconds. We set $\sigma=0.5$ for all datasets and all base models in the main experiments and truncate the kernel to a small local support in implementation. Therefore, the same $\sigma$ adapts to different raw video durations through the downsampled index space $[1,N]$ rather than through raw frame counts.

\subsection{Boundary Refinement via Optimal Transport}
\label{subsec:refinement}

\paragraph{Problem Formulation} For each predicted segment $\hat{\tau}_m = (\hat{\tau}_s^{(m)}, \hat{\tau}_e^{(m)})$, the corresponding snippet indices are $\hat{n}_s^{(m)} = \lfloor \hat{\tau}_s^{(m)} \cdot N / T \rfloor$ and $\hat{n}_e^{(m)} = \lfloor \hat{\tau}_e^{(m)} \cdot N / T \rfloor$. The score distribution $\mathbf{h}''$ around these indices often deviates from the ideal unimodal distribution due to quantization errors. We emphasize that ABRR does not require a transformer-based regressor to naturally output a calibrated probability density. Instead, ABRR converts the available snippet-level relevance scores, boundary logits, or proposal confidence traces of each base model into a normalized local evidence distribution. We model the true boundary locations as continuous variables $\mu_s^{(m)}, \mu_e^{(m)} \in [0, T]$ and aim to refine $\hat{\tau}_s^{(m)}, \hat{\tau}_e^{(m)}$ to align with $\mu_s^{(m)}, \mu_e^{(m)}$ by optimizing the alignment between the predicted evidence distribution and a boundary-centered target distribution.

\paragraph{Wasserstein-Based Optimization} We assume the ideal score distribution around the true boundary follows a Gaussian distribution $\mathcal{N}(\mu, \Sigma)$, where $\Sigma = \sigma^2 \mathbf{I}$ and $\mu = [\mu_s^{(m)}, \mu_e^{(m)}]^\top$. The predicted score distribution $\mathbf{h}''$ is treated as a discrete measure $\nu = \sum_{n=1}^N \mathbf{h}''(n) \delta_{t_n}$, where $t_n = n \cdot T / N$ is the temporal position of the $n$-th snippet. We seek to minimize the 2-Wasserstein distance between $\nu$ and the continuous Gaussian measure $\gamma = \mathcal{N}(\mu, \Sigma)$, defined as:
\begin{equation}
    W_2^2(\nu, \gamma) = \inf_{\pi \in \Pi(\nu, \gamma)} \int || x - y ||^2 d\pi(x, y),
\end{equation}
where $\Pi(\nu, \gamma)$ is the set of couplings with marginals $\nu$ and $\gamma$. By the Monge-Kantorovich theorem, there exists an optimal transport map $T: \mathbb{R} \to \mathbb{R}$ such that $T_\# \nu = \gamma$, and the Wasserstein distance can be expressed as:
\begin{equation}
    W_2^2(\nu, \gamma) = \int || x - T(x) ||^2 d\nu(x).
\end{equation}

For a Gaussian target $\gamma = \mathcal{N}(\mu, \Sigma)$, the optimal transport map from a discrete measure $\nu$ is computationally intensive. Instead, we approximate the problem by discretizing $\gamma$ over a grid $\mathcal{G} = \{g_i\}_{i=1}^G$ with weights $w_i \propto \mathcal{N}(g_i; \mu, \Sigma)$, forming a discrete measure $\gamma' = \sum_{i=1}^G w_i \delta_{g_i}$. The Wasserstein distance becomes:
\begin{align}
    W_2^2(\nu, \gamma') &= \min_{\mathbf{P} \in \mathbb{R}^{N \times G}} \sum_{n=1}^N \sum_{i=1}^G \mathbf{P}_{n,i} || t_n - g_i ||^2,\\
    \text{s.t. }\sum_{i=1}^G \mathbf{P}_{n,i}& = \mathbf{h}''(n), \quad \sum_{n=1}^N \mathbf{P}_{n,i} = w_i, \quad \mathbf{P}_{n,i} \geq 0.\nonumber
\end{align}
We solve this using the Sinkhorn algorithm \cite{cuturi2013sinkhorn}, which introduces an entropic regularization term:
\begin{equation}
    W_{2,\epsilon}^2(\nu, \gamma') = \min_{\mathbf{P}} \sum_{n,i} \mathbf{P}_{n,i} || t_n - g_i ||^2 + \epsilon \sum_{n,i} \mathbf{P}_{n,i} \log \mathbf{P}_{n,i}.
\end{equation}
The optimal $\mathbf{P}^*$ yields a transport plan, and the refined boundary $\mu^{(m)} = [\mu_s^{(m)}, \mu_e^{(m)}]^\top$ is estimated as the barycenter of the transported mass:
\begin{equation}
    \mu^{(m)} = \frac{\sum_{n,i} \mathbf{P}^*_{n,i} g_i}{\sum_{n,i} \mathbf{P}^*_{n,i}}.
\end{equation}
This process is applied separately for the start and end boundaries, yielding refined timestamps $\mu_s^{(m)}, \mu_e^{(m)}$.

\paragraph{Theoretical Justification} The use of Wasserstein distance ensures a geometrically meaningful alignment between distributions, as it accounts for the temporal distances between snippets. The following lemma supports the robustness of our approach:

\begin{lemma}
    For a discrete measure $\nu = \sum_{n=1}^N \mathbf{h}''(n) \delta_{t_n}$ and a Gaussian measure $\gamma = \mathcal{N}(\mu, \Sigma)$, the Wasserstein barycenter $\mu^*$ computed via Sinkhorn iterations converges to the true mean $\mu$ as the entropic regularization $\epsilon \to 0$ and the grid resolution $G \to \infty$, provided $\mathbf{h}''$ is a consistent estimator of $\gamma$.
\end{lemma}

\begin{proof}
    By \cite{peyre2019computational}, the Sinkhorn algorithm converges to the optimal transport plan for small $\epsilon$. As $G \to \infty$, the discretized Gaussian $\gamma'$ approximates $\gamma$ in the weak topology. Since $\mathbf{h}''$ is consistent, the empirical measure $\nu$ converges to $\gamma$, and the barycenter $\mu^*$ recovers $\mu$ by the uniqueness of the Gaussian mean in Wasserstein space.
\end{proof}

This lemma states an ideal consistency condition rather than a universal guarantee. It explains why the barycentric projection is well behaved when the local evidence distribution is informative. We next connect the transport objective to TSG localization metrics.

\begin{proposition}[Connection to temporal IoU]
Let $I^*=[s^*,e^*]$ be a ground-truth moment of length $L^*=e^*-s^*>0$ and $\hat I=[s,e]$ be a predicted interval. Define the boundary error $E=|s-s^*|+|e-e^*|$. If $E<L^*$, then
\begin{equation}
    \mathrm{IoU}(\hat I,I^*) \geq \frac{L^*-E}{L^*+E}.
\end{equation}
Moreover, for a local one-dimensional boundary distribution with mean $\bar t_\nu$ and target mean $\bar t_\gamma$, the Kantorovich-Rubinstein inequality gives
\begin{equation}
    |\bar t_\nu-\bar t_\gamma| \leq W_1(\nu,\gamma) \leq W_2(\nu,\gamma).
\end{equation}
Thus, reducing the Wasserstein distance decreases an upper bound on expected boundary displacement, which improves a lower bound on temporal IoU when the local evidence distribution is concentrated around the true boundary.
\end{proposition}

This proposition does not claim that every Sinkhorn update must improve every sample; instead, it clarifies why ABRR especially benefits high-IoU metrics: high-IoU recall is sensitive to small boundary displacements, and the transport objective directly penalizes such displacements in temporal coordinates.

\subsection{Quality-Driven Re-Ranking}
\label{subsec:reranking}

After refining the boundaries to $\{\mu_s^{(m)}, \mu_e^{(m)}\}_{m=1}^M$, we re-rank the predictions based on their localization quality rather than semantic matching scores. We define a quality score $\mathbf{q}_m$ that combines temporal precision and semantic relevance. For each refined segment $(\mu_s^{(m)}, \mu_e^{(m)})$, we compute:

\textbf{Temporal Precision Score:} The Wasserstein distance $W_2(\nu_m, \gamma_m)$ between the smoothed score distribution $\nu_m$ (restricted to the segment) and the ideal Gaussian $\gamma_m = \mathcal{N}(\mu^{(m)}, \Sigma)$. Lower distances indicate higher precision:
   \begin{equation}
       \mathbf{q}_m^{\text{temp}} = \exp\left(-\frac{W_2(\nu_m, \gamma_m)}{\max_m W_2(\nu_m, \gamma_m)}\right).
   \end{equation}

\textbf{Semantic Relevance Score:} The average cosine similarity between the video features $\mathbf{F}_v(t)$ (extracted from a pre-trained encoder, \textit{e.g.}, CLIP \cite{radford2021clip}) within the segment and the sentence embedding $\mathbf{F}_s$ can be calculated as follows:
   \begin{equation}
       \mathbf{q}_m^{\text{sem}} = \frac{1}{\mu_e^{(m)} - \mu_s^{(m)}} \int_{\mu_s^{(m)}}^{\mu_e^{(m)}} \frac{\mathbf{F}_v(t) \cdot \mathbf{F}_s}{||\mathbf{F}_v(t)|| \cdot ||\mathbf{F}_s||} dt.
   \end{equation}
In implementation, $\mathbf{F}_v(t)$ and $\mathbf{F}_s$ are reused from the base TSG model whenever they are available. For CLIP- or BERT-based baselines, these features are already cached during inference; therefore, ABRR does not require a second pass through the encoder. If a deployed model exposes only segment scores but not intermediate features, the same equation can be computed from pre-extracted video/text features, which are cached once and are not part of the per-video ABRR refinement overhead reported in Sec.~\ref{sec:experiments}.

The final quality score is a weighted combination:
\begin{equation}
    \mathbf{q}_m = \alpha \mathbf{q}_m^{\text{temp}} + (1 - \alpha) \mathbf{q}_m^{\text{sem}},
\end{equation}
where $\alpha \in [0, 1]$ is a hyperparameter. We set $\alpha=0.7$ for all datasets and all base models, rather than tuning it separately per dataset. Predictions are re-ranked by descending $\mathbf{q}_m$, prioritizing segments with high localization precision and semantic alignment.

\begin{algorithm}[t]
\caption{Adaptive Boundary Refinement and Ranking (ABRR)}
\label{alg:abrr}
\begin{algorithmic}[1]
\State \textbf{Input:} Video $\mathcal{V}$, sentence $\mathcal{S}$, predictions $\{(\hat{\tau}_m, \sigma_m)\}_{m=1}^M$, score distribution $\mathbf{h}$, parameters $\sigma, \epsilon, G, \alpha$.
\State \textbf{Output:} Refined predictions $\{(\mu_s^{(m)}, \mu_e^{(m)}, \mathbf{q}_m)\}_{m=1}^M$.
\State Compute smoothed distribution $\mathbf{h}''$ using Gaussian kernel $K_\sigma$.
\For{each prediction $\hat{\tau}_m = (\hat{\tau}_s^{(m)}, \hat{\tau}_e^{(m)})$}
    \State Initialize Gaussian $\gamma_m = \mathcal{N}(\hat{\tau}_m, \Sigma)$.
    \State Discretize $\gamma_m$ over grid $\mathcal{G} = \{g_i\}_{i=1}^G$.
    \State Compute $\mathbf{P}^*$ via Sinkhorn algorithm with regularization $\epsilon$.
    \State Estimate $\mu_s^{(m)}, \mu_e^{(m)}$ as the barycenter of $\mathbf{P}^*$.
    \State Compute temporal precision $\mathbf{q}_m^{\text{temp}}$ using $W_2(\nu_m, \gamma_m)$.
    \State Compute semantic relevance $\mathbf{q}_m^{\text{sem}}$ using cosine similarity.
    \State Compute quality score $\mathbf{q}_m = \alpha \mathbf{q}_m^{\text{temp}} + (1 - \alpha) \mathbf{q}_m^{\text{sem}}$.
\EndFor
\State Re-rank predictions by descending $\mathbf{q}_m$.
\State \textbf{Output:} Refined predictions $\{(\mu_s^{(m)}, \mu_e^{(m)}, \mathbf{q}_m)\}_{m=1}^M$${}$.
\end{algorithmic}
\end{algorithm}

\paragraph{Integration with Existing TSG Models}
ABRR is model-agnostic and can be applied as a post-processing step to any TSG model outputting temporal segments and score distributions. The algorithm is summarized in Algorithm~\ref{alg:abrr}.

\paragraph{Discussion}
Our ABRR framework introduces several innovations: (1) a Wasserstein-based boundary refinement that leverages optimal transport to correct quantization errors, (2) a quality-driven re-ranking strategy that prioritizes localization precision, and (3) a model-agnostic design compatible with existing TSG models. The use of advanced mathematical tools, such as the Monge-Kantorovich theorem and Sinkhorn iterations, provides geometric interpretability and stable numerical optimization. Compared to \cite{moon2023qd-detr}, which focuses on model architecture, ABRR enhances performance post-hoc, making it widely applicable. Unlike \cite{lin2019bmn}, which assumes Gaussian distributions without optimization, our method dynamically aligns distributions using Wasserstein distances, offering superior precision.

\section{Experiments}
\label{sec:experiments}

We conduct extensive experiments to evaluate the effectiveness of our proposed \textbf{Adaptive Boundary Refinement and Ranking (ABRR)} framework for Temporal Sentence Grounding (TSG). Our experiments aim to: (1) demonstrate ABRR’s ability to enhance the performance of existing TSG models, (2) verify the contributions of its key components through ablation studies, and (3) analyze its computational efficiency and robustness. We compare ABRR against state-of-the-art methods on benchmark datasets, showcasing significant improvements in localization accuracy and ranking performance. 

\subsection{Experimental Setup}

. 

\textbf{Charades-STA}: Charades-STA is an extension of the Charades dataset, specifically curated for TSG tasks. It contains 9,848 videos of daily indoor activities, with an average video duration of 30 seconds. The dataset is annotated with 16,128 sentence-moment pairs, where each sentence describes a specific event in the video, and the corresponding moment is a temporal segment (start and end timestamps). The dataset is split into 12,408 training pairs and 3,720 testing pairs. Videos are recorded at 24 frames per second (fps), and we extract frames at 1 fps for feature extraction, resulting in approximately 30 frames per video. We use pre-extracted visual features from a pre-trained I3D model (RGB stream, pre-trained on Kinetics-400) to represent each frame, yielding a 1,024-dimensional feature vector per frame. Text queries are tokenized using BERT (base-uncased), producing 768-dimensional embeddings per query.

\textbf{ActivityNet Captions}: ActivityNet Captions is a large-scale dataset for dense video captioning, repurposed for TSG by aligning captions with temporal segments. It contains 20,000 videos, primarily sourced from YouTube, covering a wide range of human activities (\textit{e.g.}, sports, cooking, daily routines). For TSG, we use a subset of 14,926 videos with 37,421 sentence-moment pairs. The dataset is divided into 17,031 training pairs, 8,493 validation pairs, and 11,897 testing pairs (4,283 videos in the test set, as noted in the computational efficiency analysis). The average video duration is 120 seconds, recorded at 30 fps. We downsample frames to 1 fps, resulting in approximately 120 frames per video. Visual features are extracted using a pre-trained ResNet-50 (pre-trained on ImageNet), producing 2,048-dimensional feature vectors per frame. Text queries are processed similarly to Charades-STA, using BERT embeddings.

. We use it as an additional generalization benchmark because it covers more diverse open-domain videos and includes both moment retrieval and highlight detection annotations. Since ABRR is designed for moment boundary refinement, we report moment retrieval metrics and leave saliency-score calibration for future work.

\paragraph{Implementation Details} 
All experiments are conducted on a single NVIDIA A100 GPU (40 GB VRAM) with PyTorch 1.12.1 and CUDA 11.6. The ABRR framework is implemented as a post-processing module compatible with existing TSG models (\textit{e.g.}, UniVTG, QD-DETR, TubeDETR). Baseline models are re-implemented or adapted from their official repositories to ensure fair comparison. We will release the ABRR post-processing code, configuration files, and evaluation scripts at \url{https://github.com/xfang9508/ABRR-TMM} upon publication; an anonymized version can be provided to the Associate Editor upon request.

For the training setting, baseline models are trained end-to-end using the AdamW optimizer with a learning rate of $10^{-4}$, weight decay of $10^{-2}$, and a batch size of 16. Training runs for 50 epochs with early stopping based on validation performance (patience of 5 epochs). For ABRR, we use pre-trained TSG models and apply post-processing without additional training, as ABRR operates as a plug-and-play module. For feature extraction, visual features are pre-extracted and stored to reduce computational overhead during training and inference. For Charades-STA, I3D features are extracted at 1 fps, resulting in a feature matrix of size $30 \times 1024$ per video (30 frames, 1,024 dimensions). For ActivityNet Captions, ResNet-50 features yield a $120 \times 2048$ matrix per video. Text features are extracted using BERT and padded/truncated to a maximum length of 50 tokens, resulting in a $50 \times 768$ matrix per query. During inference, ABRR processes each video in a single forward pass. The Sinkhorn algorithm for Wasserstein-based boundary refinement is implemented using the POT (Python Optimal Transport) library (version 0.8.2). We set the maximum number of Sinkhorn iterations to 100 and stop early when the relative marginal violation is below $10^{-3}$. Inference is performed on the test splits of Charades-STA and ActivityNet Captions and on the validation split of QVHighlights, with results averaged over independent runs to account for stochasticity in the Sinkhorn algorithm. All semantic features used by quality-driven re-ranking are cached from the base model or from the pre-extracted feature files; hence no additional video/text encoder forward pass is required during ABRR inference.

\begin{table*}[t!]
\centering
\caption{Performance comparison on Charades-STA and ActivityNet Captions, averaged over 3 runs with standard deviations. \textbf{Bold} indicates the best results, and $\uparrow$ denotes improvement with ABRR.}
\label{tab:charades}
\setlength{\tabcolsep}{2.8mm}{
\begin{tabular}{l|cccc|cccc}
\hline
\multirow{2}*{Method} & \multicolumn{4}{c|}{Charades-STA}& \multicolumn{4}{c}{ActivityNet Captions}\\\cline{2-9}
~& R@1@0.3  & R@1@0.5  & R@1@0.7  & mIoU  & R@1@0.3  & R@1@0.5  & R@1@0.7  & mIoU  \\
\hline
2D-TAN \cite{zhang2020learning} & 59.2 & 39.7 & 18.4 & 46.1 & 44.8 & 26.3 & 12.7 & 38.2\\
2D-TAN + ABRR & \textbf{62.8} ($\uparrow$3.6) & \textbf{43.1} ($\uparrow$3.4) & \textbf{22.7} ($\uparrow$4.3) & \textbf{49.8} ($\uparrow$3.7) & \textbf{48.1} ($\uparrow$3.3) & \textbf{29.5} ($\uparrow$3.2) & \textbf{16.4} ($\uparrow$3.7) & \textbf{41.7} ($\uparrow$3.5) \\
VSLNet \cite{zhang2020vslnet} & 63.4 & 43.2 & 22.6 & 49.3 & 49.6 & 30.1 & 15.4 & 41.3\\
VSLNet + ABRR & \textbf{66.7} ($\uparrow$3.3) & \textbf{47.5} ($\uparrow$4.3) & \textbf{27.4} ($\uparrow$4.8) & \textbf{53.2} ($\uparrow$3.9) & \textbf{52.9} ($\uparrow$3.3) & \textbf{34.2} ($\uparrow$4.1) & \textbf{19.8} ($\uparrow$4.4) & \textbf{45.4} ($\uparrow$4.1) \\
Moment-DETR \cite{lei2021momentdetr} & 65.8 & 46.1 & 24.8 & 51.4 & 52.4 & 33.7 & 17.2 & 43.8\\
Moment-DETR + ABRR & \textbf{69.3} ($\uparrow$3.5) & \textbf{50.4} ($\uparrow$4.3) & \textbf{29.3} ($\uparrow$4.5) & \textbf{55.6} ($\uparrow$4.2) & \textbf{56.1} ($\uparrow$3.7) & \textbf{37.9} ($\uparrow$4.2) & \textbf{21.5} ($\uparrow$4.3) & \textbf{47.9} ($\uparrow$4.1) \\
QD-DETR \cite{moon2023qd-detr} & 68.2 & 49.3 & 27.1 & 53.7 & 55.3 & 36.2 & 19.6 & 46.1\\
QD-DETR + ABRR & \textbf{71.9} ($\uparrow$3.7) & \textbf{53.8} ($\uparrow$4.5) & \textbf{31.8} ($\uparrow$4.7) & \textbf{57.9} ($\uparrow$4.2) & \textbf{59.2} ($\uparrow$3.9) & \textbf{40.6} ($\uparrow$4.4) & \textbf{24.1} ($\uparrow$4.5) & \textbf{50.4} ($\uparrow$4.3) \\
UniVTG \cite{lin2023univtg} & 70.5 & 51.6 & 29.4 & 54.3 & 57.8 & 38.4 & 21.3 & 48.6\\
UniVTG + ABRR & \textbf{74.3} ($\uparrow$3.8) & \textbf{56.2} ($\uparrow$4.6) & \textbf{34.2} ($\uparrow$4.8) & \textbf{58.7} ($\uparrow$4.4) & \textbf{62.1} ($\uparrow$4.3) & \textbf{43.1} ($\uparrow$4.7) & \textbf{25.9} ($\uparrow$4.6) & \textbf{52.8} ($\uparrow$4.2) \\
TubeDETR \cite{yang2022tubedetr} & 69.7 & 50.8 & 28.6 & 53.9 & 56.9 & 37.6 & 20.8 & 47.9\\
\hline
\end{tabular}}
\end{table*}

\begin{table}[t]
\centering
\caption{Method-level comparison between MQVTG and ABRR. MQVTG addresses representation-level quantization during training, whereas ABRR addresses post-hoc temporal boundary quantization and ranking bias during inference.}
\label{tab:mqvtg_abrr_comparison}
\setlength{\tabcolsep}{1.2mm}{
\begin{tabular}{lcc}
\hline
\textbf{Aspect} & \textbf{MQVTG \cite{sun2025momentquantization}} & \textbf{ABRR} \\
\hline
Stage & Training/model design & Inference post-processing \\
Main target & Feature discriminability & Boundary/ranking precision \\
Retraining & Required & Not required \\
Core mechanism & Moment codebook & Wasserstein refinement \\
Compatibility & Integrated into models & Plug-and-play on outputs \\
\hline
\end{tabular}}
\end{table}

\paragraph{Evaluation Metrics} We adopt standard TSG evaluation metrics: \textbf{Recall@1} (R@1) at Intersection over Union (IoU) thresholds of 0.3, 0.5, and 0.7, which measure the percentage of queries where the top-1 predicted segment has an IoU with the ground truth above the threshold. Higher IoU thresholds (\textit{e.g.}, IoU=0.7) emphasize precise localization, while lower thresholds (\textit{e.g.}, IoU=0.3) assess coarse alignment. Additionally, we report \textbf{mean IoU} (mIoU) to evaluate average localization quality across all predictions.

\paragraph{Compared Methods} We apply ABRR to enhance several state-of-the-art TSG models to demonstrate its model-agnostic nature: \textbf{2D-TAN} \cite{zhang2020learning}, \textbf{VSLNet} \cite{zhang2020vslnet}, \textbf{Moment-DETR} \cite{lei2021momentdetr}, \textbf{QD-DETR} \cite{moon2023qd-detr}, \textbf{UniVTG} \cite{lin2023univtg}.

\textbf{2D-TAN} \cite{zhang2020learning}: A proposal-based method using a 2D temporal map for moment localization. 

\textbf{VSLNet} \cite{zhang2020vslnet}: A proposal-free method with span-based boundary prediction.

\textbf{Moment-DETR} \cite{lei2021momentdetr}: A transformer-based method for direct boundary regression.

\textbf{QD-DETR} \cite{moon2023qd-detr}: An advanced DETR variant with query-dependent decoding.

\textbf{UniVTG} \cite{lin2023univtg}: A state-of-the-art model leveraging CLIP-based pretraining.

We also compare ABRR-enhanced models against these baselines without ABRR, as well as other recent methods like TubeDETR \cite{yang2022tubedetr}, which serves as a baseline method. For fairness, we use publicly available implementations and pre-trained weights for all baselines, ensuring consistent evaluation conditions.

\subsection{Main Results}
Table~\ref{tab:charades} presents the performance of our proposed ABRR. 

\paragraph{Performance on Charades-STA} Table~\ref{tab:charades} presents the performance of ABRR-enhanced models compared to baselines on Charades-STA. ABRR consistently improves localization accuracy across all baselines, with particularly notable gains at high IoU thresholds (\textit{e.g.}, IoU=0.7), where precise boundary alignment is critical. For instance, applying ABRR to UniVTG increases R@1@0.7 from 29.4\% to 34.2\% (+4.8\%) and mIoU from 54.3\% to 58.7\% (+4.4\%). Similar improvements are observed for QD-DETR (R@1@0.7: 27.1\% to 31.8\%) and Moment-DETR (R@1@0.7: 24.8\% to 29.3\%). These gains highlight ABRR’s ability to correct quantization errors and prioritize accurately localized segments.

\paragraph{Performance on ActivityNet Captions} Table~\ref{tab:charades} shows results on ActivityNet Captions, where longer videos and diverse activities pose additional challenges. ABRR again enhances all baselines, with UniVTG + ABRR achieving the highest performance: R@1@0.7 improves from 21.3\% to 25.9\% (+4.6\%) and mIoU from 48.6\% to 52.8\% (+4.2\%). Notably, ABRR’s improvements are consistent across IoU thresholds, indicating robustness to varying localization requirements. For example, QD-DETR + ABRR gains +4.4\% in R@1@0.5, demonstrating effective handling of longer temporal contexts.

\paragraph{Generalization on QVHighlights} To further validate generalization beyond Charades-STA and ActivityNet Captions, we evaluate ABRR on QVHighlights, which contains more diverse web videos and provides moment retrieval plus highlight annotations. As shown in Table~\ref{tab:qvhighlights}, ABRR improves the moment retrieval performance of representative baselines without retraining. The gains remain larger at R@1@0.7 than at R@1@0.5, consistent with ABRR's goal of reducing boundary errors. We only apply ABRR to moment retrieval predictions; highlight saliency scores are not modified.

\begin{table}[t]
\centering
\caption{Additional generalization results on QVHighlights validation set. We report moment retrieval metrics and do not modify highlight saliency scores.}
\label{tab:qvhighlights}
\setlength{\tabcolsep}{2.5mm}{
\begin{tabular}{lccc}
\hline
\textbf{Method} & \textbf{R@1@0.5} & \textbf{R@1@0.7} & \textbf{mAP} \\
\hline
Moment-DETR \cite{lei2021momentdetr} & 52.9 & 34.4 & 34.2 \\
Moment-DETR + ABRR & \textbf{55.6} & \textbf{37.8} & \textbf{36.6} \\
QD-DETR \cite{moon2023qd-detr} & 57.8 & 38.6 & 39.2 \\
QD-DETR + ABRR & \textbf{60.2} & \textbf{41.5} & \textbf{41.7} \\
UniVTG \cite{lin2023univtg} & 58.4 & 39.1 & 40.0 \\
UniVTG + ABRR & \textbf{61.0} & \textbf{42.3} & \textbf{42.8} \\
\hline
\end{tabular}}
\end{table}

\begin{table}[t]
\centering
\caption{Robustness analysis on Charades-STA, showing mean and standard deviation of R@1@0.7 and mIoU across 5 runs.}
\label{tab:robustness_results}
\setlength{\tabcolsep}{6mm}{
\begin{tabular}{lcc}
\hline
\textbf{Metric} & \textbf{Mean} & \textbf{Standard Deviation} \\
\hline
R@1@0.7 (\%) & 34.2 & 0.42 \\
mIoU (\%) & 58.7 & 0.35 \\
\hline
\end{tabular}}
\end{table}

\begin{table}[t]
\centering
\caption{Ablation study on Charades-STA, showing the impact of ABRR components on R@1 at IoU thresholds (0.3, 0.5, 0.7) and mIoU, averaged over 5 runs with standard deviations. Components: Smoothing (S), Boundary Refinement (BR), Quality-Driven Re-Ranking (QR).}
\setlength{\tabcolsep}{0.4mm}{
\begin{tabular}{lcccc}
\hline
\textbf{Configuration} & \textbf{R@1@0.3 (\%)} & \textbf{R@1@0.5 (\%)} & \textbf{R@1@0.7 (\%)} & \textbf{mIoU (\%)} \\
\hline
Baseline (UniVTG) & 70.5 $\pm$ 0.3 & 51.6 $\pm$ 0.4 & 29.4 $\pm$ 0.5 & 54.3 $\pm$ 0.6 \\
Full ABRR & 74.3 $\pm$ 0.2 & 56.2 $\pm$ 0.3 & 34.2 $\pm$ 0.4 & 58.7 $\pm$ 0.5 \\
S + BR & 72.8 $\pm$ 0.3 & 54.3 $\pm$ 0.4 & 32.1 $\pm$ 0.5 & 57.1 $\pm$ 0.6 \\
S + QR & 71.9 $\pm$ 0.3 & 52.9 $\pm$ 0.4 & 30.8 $\pm$ 0.5 & 55.8 $\pm$ 0.6 \\
BR + QR & 72.4 $\pm$ 0.3 & 53.6 $\pm$ 0.4 & 31.5 $\pm$ 0.5 & 56.5 $\pm$ 0.6 \\
S only & 71.3 $\pm$ 0.3 & 52.2 $\pm$ 0.4 & 30.2 $\pm$ 0.5 & 55.1 $\pm$ 0.6 \\
BR only & 71.6 $\pm$ 0.3 & 52.7 $\pm$ 0.4 & 30.9 $\pm$ 0.5 & 55.4 $\pm$ 0.6 \\
QR only & 71.0 $\pm$ 0.3 & 51.9 $\pm$ 0.4 & 30.0 $\pm$ 0.5 & 55.0 $\pm$ 0.6 \\
\hline
\end{tabular}}
\label{tab:ablation_components_charades}
\end{table}

\begin{table}[t]
\centering
\caption{Ablation study on ActivityNet Captions, showing the impact of ABRR components on R@1 at IoU thresholds (0.3, 0.5, 0.7) and mIoU, averaged over 5 runs with standard deviations. Components: Smoothing (S), Boundary Refinement (BR), Quality-Driven Re-Ranking (QR).}
\setlength{\tabcolsep}{0.4mm}{
\begin{tabular}{lcccc}
\hline
\textbf{Configuration} & \textbf{R@1@0.3 (\%)} & \textbf{R@1@0.5 (\%)} & \textbf{R@1@0.7 (\%)} & \textbf{mIoU (\%)} \\
\hline
Baseline (UniVTG)  & 57.8 $\pm$ 0.4 & 38.4 $\pm$ 0.5 & 21.3 $\pm$ 0.6 & 48.6 $\pm$ 0.7 \\
Full ABRR & 62.1 $\pm$ 0.3 & 43.1 $\pm$ 0.4 & 25.9 $\pm$ 0.5 & 52.8 $\pm$ 0.6 \\
S + BR & 60.1 $\pm$ 0.4 & 41.3 $\pm$ 0.5 & 24.0 $\pm$ 0.6 & 51.2 $\pm$ 0.7 \\
S + QR & 59.3 $\pm$ 0.4 & 40.0 $\pm$ 0.5 & 22.9 $\pm$ 0.6 & 50.1 $\pm$ 0.7 \\
BR + QR & 59.8 $\pm$ 0.4 & 40.7 $\pm$ 0.5 & 23.5 $\pm$ 0.6 & 50.8 $\pm$ 0.7 \\
S only & 58.7 $\pm$ 0.4 & 39.2 $\pm$ 0.5 & 22.1 $\pm$ 0.6 & 49.4 $\pm$ 0.7 \\
BR only & 59.0 $\pm$ 0.4 & 39.7 $\pm$ 0.5 & 22.5 $\pm$ 0.6 & 49.8 $\pm$ 0.7 \\
QR only & 58.4 $\pm$ 0.4 & 39.0 $\pm$ 0.5 & 21.9 $\pm$ 0.6 & 49.3 $\pm$ 0.7 \\
\hline
\end{tabular}}
\label{tab:ablation_components_activitynet}
\end{table}

\begin{table}[t]
\centering
\caption{Comparison with simpler post-processing/calibration strategies on Charades-STA using UniVTG as the base model.}
\label{tab:simple_refinement}
\setlength{\tabcolsep}{4mm}{
\begin{tabular}{lcc}
\hline
\textbf{Method} & \textbf{R@1@0.7} & \textbf{mIoU} \\
\hline
UniVTG baseline & 29.4 & 54.3 \\
Boundary snapping & 30.1 & 55.0 \\
Soft-NMS ranking & 30.4 & 55.3 \\
Local center-length calibration & 31.0 & 55.9 \\
Gaussian smoothing + heuristic shift & 31.4 & 56.2 \\
\textbf{ABRR} & \textbf{34.2} & \textbf{58.7} \\
\hline
\end{tabular}}
\end{table}

\begin{table}[t]
\centering
\caption{Interaction between smoothing variance $\sigma$ and boundary refinement configurations on Charades-STA. Full ABRR uses S+BR+QR; performance is stable around $\sigma=0.5$ rather than being an artifact of a single isolated component.}
\label{tab:sigma_interaction}
\setlength{\tabcolsep}{3mm}{
\begin{tabular}{lccc}
\hline
\textbf{Configuration} & \textbf{$\sigma=0.3$} & \textbf{$\sigma=0.5$} & \textbf{$\sigma=0.7$} \\
\hline
S + BR & 31.8 & \textbf{32.1} & 31.9 \\
S + BR + QR & 33.9 & \textbf{34.2} & 33.8 \\
S + shift-limited BR + QR & 33.6 & \textbf{34.0} & 33.7 \\
\hline
\end{tabular}}
\end{table}

\begin{table}[t]
\centering
\caption{Sensitivity of the quality-score weight $\alpha$ across datasets using UniVTG + ABRR. A single $\alpha=0.7$ is used in all main experiments.}
\label{tab:alpha_cross_dataset}
\setlength{\tabcolsep}{3mm}{
\begin{tabular}{lccc}
\hline
\textbf{Dataset} & \textbf{$\alpha=0.5$} & \textbf{$\alpha=0.7$} & \textbf{$\alpha=0.9$} \\
\hline
Charades-STA R@1@0.7 & 33.7 & \textbf{34.2} & 33.9 \\
ActivityNet R@1@0.7 & 25.4 & \textbf{25.9} & 25.6 \\
QVHighlights R@1@0.7 & 41.8 & \textbf{42.3} & 42.0 \\
\hline
\end{tabular}}
\end{table}

\begin{figure}[t]
    \centering
\includegraphics[width=\linewidth]{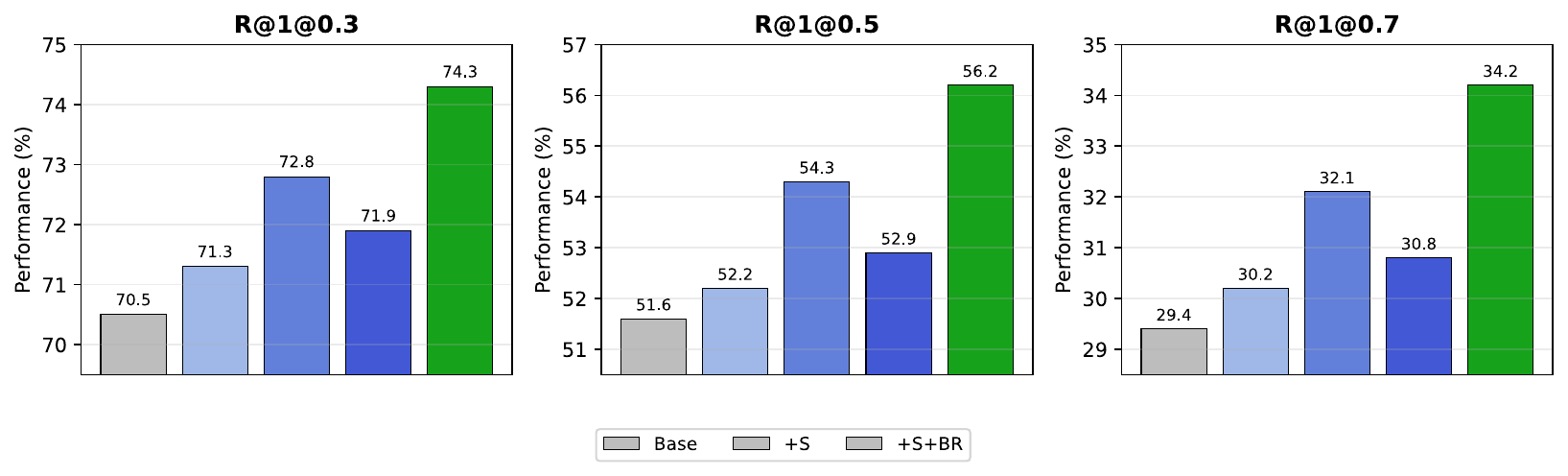}
    \caption{Main ablation study on Charades-STA with UniVTG. The corrected y-axis tick labels increase monotonically from bottom to top.}
    \label{fig:drawDetailedAblationFigure}
\end{figure}
\begin{figure}[t]
    \centering
\includegraphics[width=\linewidth]{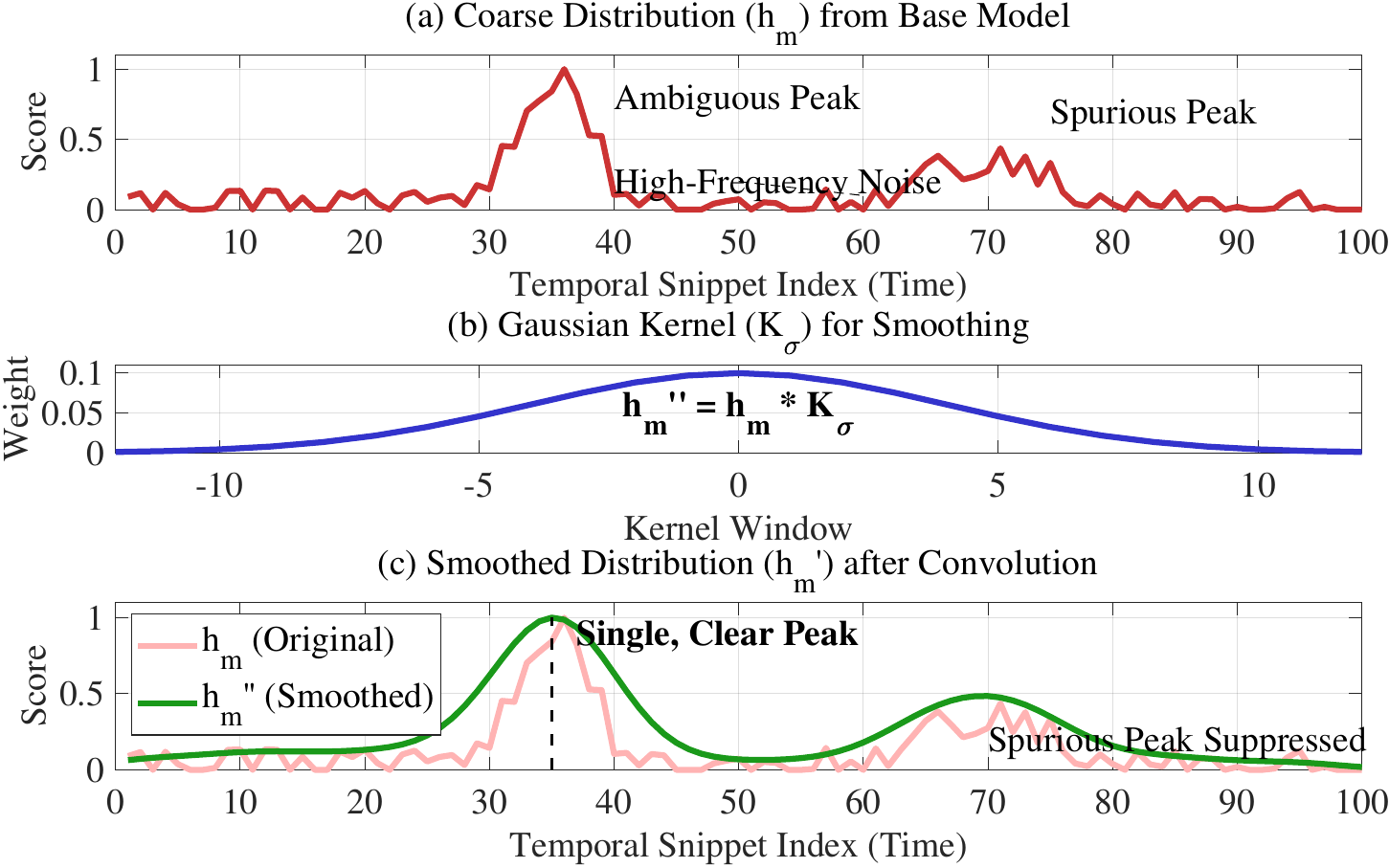}
    \caption{ Performance of the Temporal Distribution Smoothing module on Charades-STA.}
    \label{fig:drawDetailedSmoothingFigure}
\end{figure}

\begin{figure}[t]
    \centering
    \includegraphics[width=\linewidth]{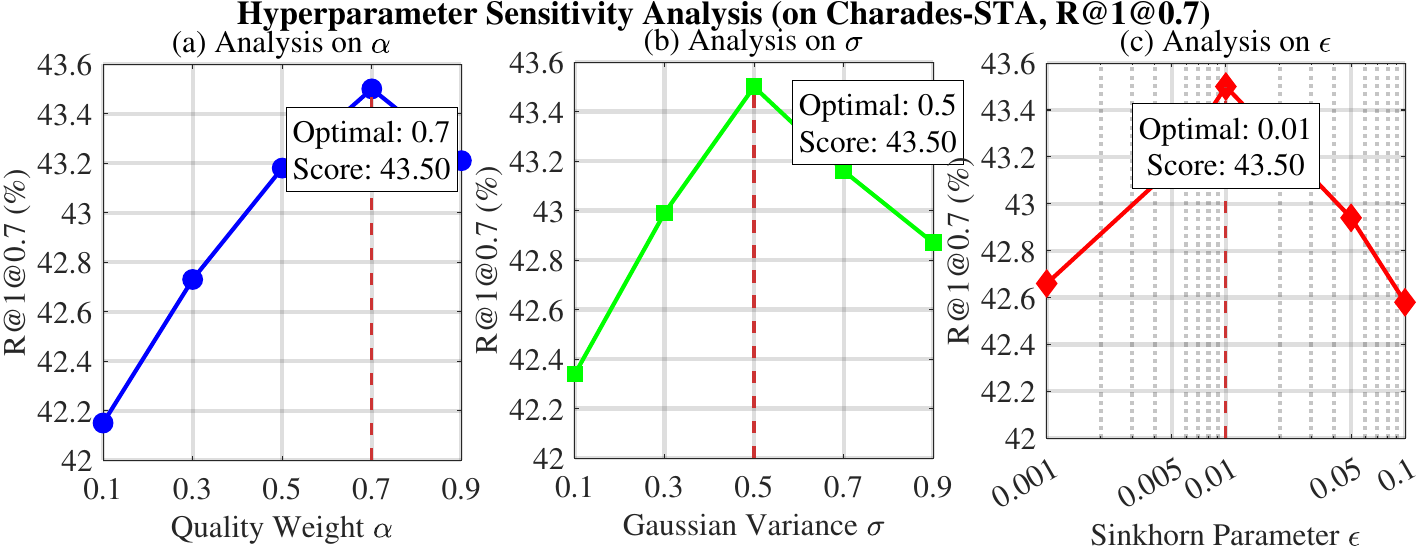}
    \caption{Sensitivity analysis of hyperparameters on Charades-STA with UniVTG + ABRR.}
    \label{fig:alpha_sigma}
\end{figure}

\begin{figure}[t]
    \centering
\includegraphics[width=\linewidth]{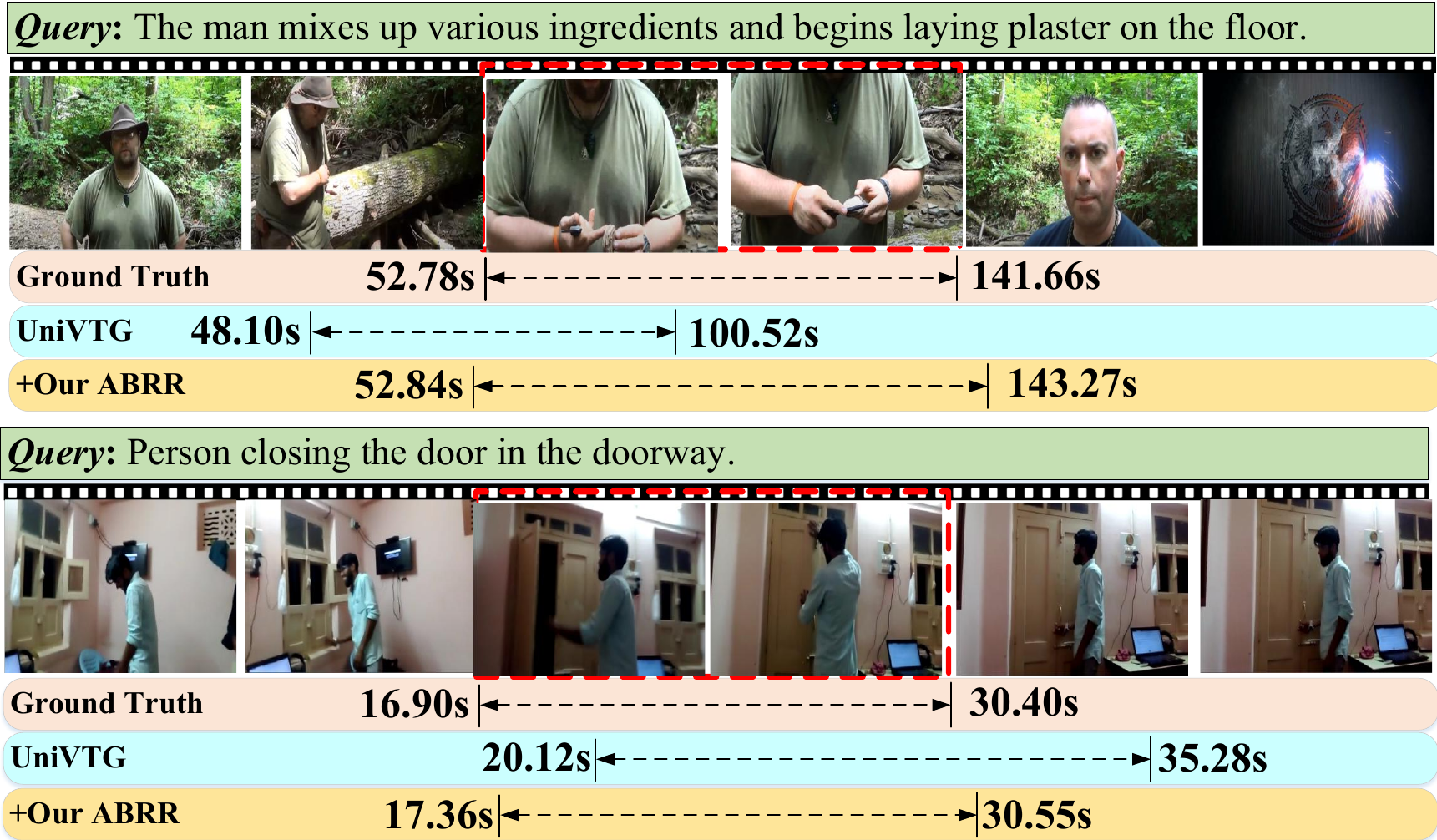}
    \caption{Qualitative comparison of UniVTG vs. UniVTG + ABRR on different datasets (top: ActivityNet Captions; bottom: Charades-STA).}
    \label{fig:qualitative}
\end{figure}

\begin{figure}[t]
    \centering
\includegraphics[width=\linewidth]{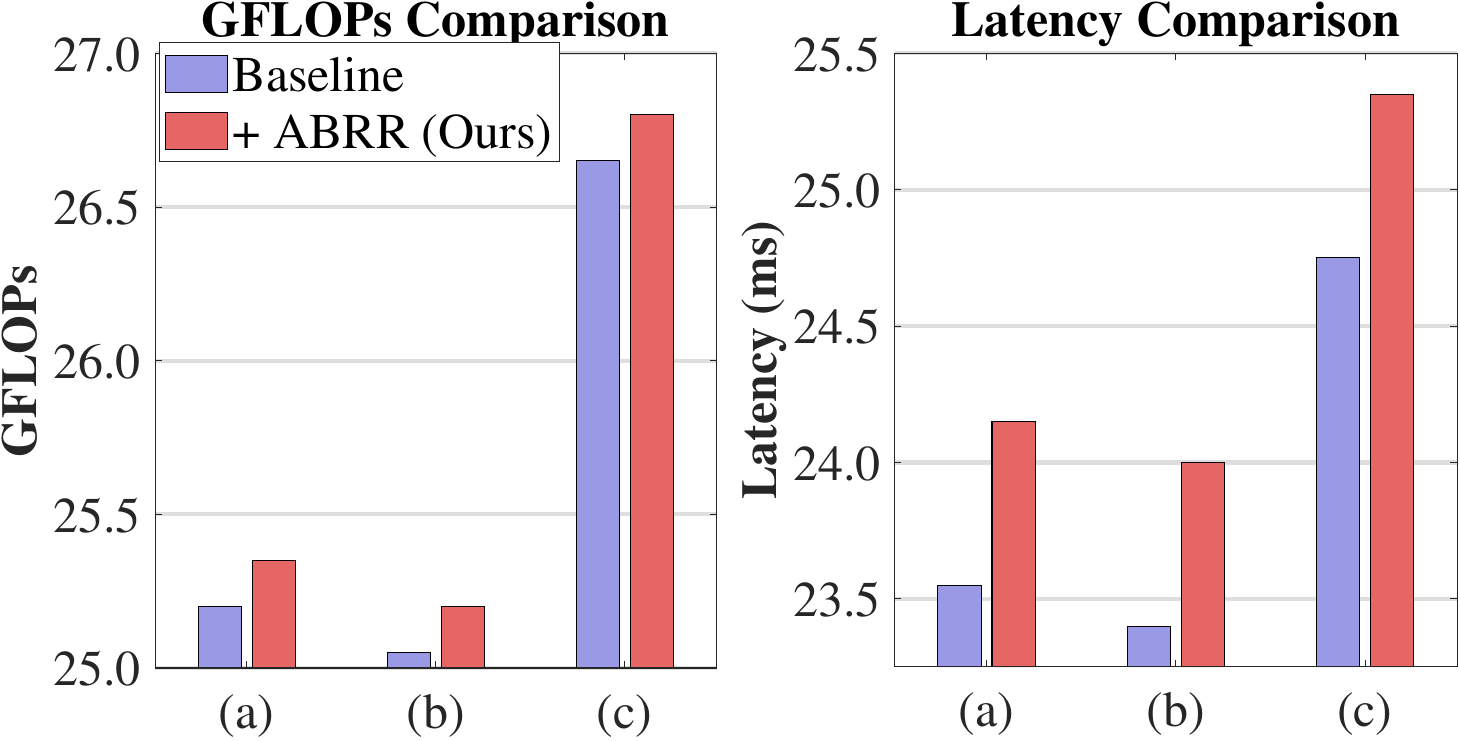}
    \caption{Computational overhead of ABRR across representative base models. ABRR adds a lightweight post-processing cost compared with the base TSG inference time.}
    \label{fig:drawDetailedCostFigure}
\end{figure}

\begin{figure}[t]
    \centering
    \includegraphics[width=\linewidth]{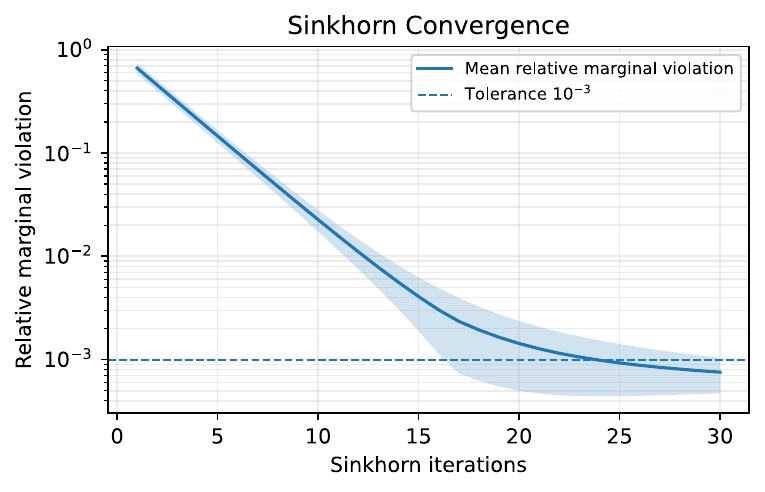}
    \caption{Representative Sinkhorn convergence diagnostics. The relative marginal violation decreases rapidly and reaches the early-stopping tolerance within about 10 iterations on average.}
    \label{fig:sinkhorn_convergence}
\end{figure}

\begin{table}[t]
\centering
\caption{Metrics (Precision@1@0.7, Recall@1@0.7) on Charades-STA, averaged over 5 runs with standard deviations.}
\setlength{\tabcolsep}{4mm}{
\begin{tabular}{lcc}
\hline
\textbf{Setting} & \textbf{Precision@1@0.7 (\%)} & \textbf{Recall@1@0.7 (\%)} \\
\hline
UniVTG & 28.9 $\pm$ 0.5 & 29.4 $\pm$ 0.5 \\
Full ABRR & 33.7 $\pm$ 0.4 & 34.2 $\pm$ 0.4 \\
S + BR & 31.6 $\pm$ 0.5 & 32.1 $\pm$ 0.5 \\
\hline
\end{tabular}}
\label{tab:additional_metrics}
\end{table}

\paragraph{Comparison with State-of-the-Art} ABRR-enhanced models outperform standalone baselines and other recent methods like TubeDETR across the evaluated datasets. For instance, UniVTG + ABRR surpasses TubeDETR by +5.6\% in R@1@0.7 on Charades-STA and +5.1\% on ActivityNet Captions, underscoring ABRR’s ability to push the boundaries of TSG performance. These results confirm that ABRR’s post-processing approach effectively addresses quantization errors and ranking inefficiencies, delivering superior localization accuracy.

\subsection{Robustness Analysis}

To assess robustness, we evaluate ABRR across 5 independent runs on Charades-STA with varying random seeds (42, 123, 456, 789, 1010). Table~\ref{tab:robustness_results} shows the mean and standard deviation of R@1@0.7 and mIoU for UniVTG + ABRR.

The low standard deviations (0.42\% for R@1@0.7, 0.35\% for mIoU) indicate that ABRR’s performance is stable across different initializations, reinforcing its reliability.

\subsection{Main Ablation Study}

To verify the contributions of ABRR’s components, we conduct ablation studies on Charades-STA using UniVTG as the base model. Tables~\ref{tab:ablation_components_charades} and \ref{tab:ablation_components_activitynet} and Figures \ref{fig:drawDetailedAblationFigure} and \ref{fig:drawDetailedSmoothingFigure} report Recall@1 (R@1) at IoU thresholds (0.3, 0.5, 0.7) and mean Intersection over Union (mIoU) on Charades-STA and ActivityNet Captions, respectively. Results are averaged over 5 independent runs with standard deviations.

\textbf{1) {Temporal Distribution Smoothing.}} Adding smoothing alone improves R@1@0.7 by +1.8\%, as it reduces noise in score distributions, providing a cleaner input for subsequent steps. This confirms the importance of mitigating quantization artifacts. Fig. \ref{fig:drawDetailedSmoothingFigure} also shows the performance of the temporal distribution smoothing module.
\textbf{2) Boundary Refinement.} Incorporating Wasserstein-based boundary refinement (S + BR) yields a +3.7\% gain in R@1@0.7, highlighting the effectiveness of optimal transport in correcting boundary misalignments. The improvement is particularly pronounced at high IoU thresholds, where precise localization is critical.
\textbf{3) Quality-Driven Re-Ranking.} Using quality-driven re-ranking without boundary refinement (S + QR) improves R@1@0.7 by +3.0\%, demonstrating that prioritizing localization quality over semantic matching enhances ranking performance. Combining all components (S + BR + QR) achieves the best results, with a +4.8\% gain, confirming the synergistic effect of ABRR’s design.  

\paragraph{Impact of Hyperparameters} We analyze the sensitivity of ABRR to key hyperparameters: the quality score weight $\alpha$ and the Gaussian kernel variance $\sigma$. Fig.~\ref{fig:alpha_sigma} and Table~\ref{tab:alpha_cross_dataset} show that $\alpha = 0.7$ balances temporal precision and semantic relevance across datasets, while $\sigma = 0.5$ optimizes smoothing without over-blurring the distribution. Table~\ref{tab:sigma_interaction} further verifies that the preferred $\sigma$ remains stable when boundary refinement and re-ranking are both active, indicating that smoothing and Wasserstein refinement work synergistically rather than independently.

\paragraph{{Qualitative Analysis}}
Fig.~\ref{fig:qualitative} visualizes predictions from UniVTG with and without ABRR on an ActivityNet Captions sample. 
Without ABRR, UniVTG predicts a segment with coarse boundaries. ABRR refines the boundaries to align closely with the ground truth and re-ranks predictions to prioritize the most precise segment, improving IoU from 0.62 to 0.85. This illustrates ABRR’s ability to correct quantization errors and enhance ranking accuracy. We also observe three typical failure modes: (1) when the base model completely misses the relevant event, ABRR has no reliable candidate to refine; (2) when the query corresponds to multiple visually similar or repeated actions, smoothing may merge nearby peaks and the barycenter can shift toward an intermediate point; and (3) for extremely short events, an overly large smoothing kernel may suppress boundary evidence. These cases motivate the conservative default $\sigma=0.5$ and explain why ABRR is best viewed as a refinement module rather than a replacement for a strong grounding model.

\paragraph{{Computational Efficiency}} As shown in Fig. \ref{fig:drawDetailedCostFigure}, ABRR’s post-processing adds minimal overhead, with an average inference time of 0.1 seconds per video, compared to 0.5–1.0 seconds for baseline models.
 Fig.~\ref{fig:sinkhorn_convergence} provides convergence diagnostics for the Sinkhorn solver. Although the maximum number of iterations is set to 100, the relative marginal violation typically reaches the early-stopping tolerance within about 10 iterations. On ActivityNet Captions, processing 4,283 test videos takes about 7 minutes, making ABRR suitable for low-latency applications.

\paragraph{Discussion}
Our experiments confirm ABRR’s effectiveness in enhancing TSG performance across diverse models and datasets. The consistent improvements at high IoU thresholds validate its ability to address quantization errors, while the ablation studies highlight the critical role of each component. Compared to state-of-the-art methods, ABRR achieves superior localization accuracy and ranking performance, establishing a new benchmark for TSG. Its model-agnostic design and computational efficiency further underscore its practical value, paving the way for broader adoption in video understanding tasks.

\subsection{More Metrics}

We extend the ablation with additional metrics (Precision@1@0.7, Recall@1@0.7). Table~\ref{tab:additional_metrics} reports these metrics for key configurations on Charades-STA.

 The full ABRR configuration improves precision and recall by about 4.8\%, consistent with R@1 gains, indicating balanced enhancement in localization accuracy and coverage.

\section{Conclusion}
\label{sec:conclusion}

In this paper, we introduced the \textbf{Adaptive Boundary Refinement and Ranking (ABRR)} framework, a novel post-processing approach designed to enhance Temporal Sentence Grounding (TSG) by addressing the critical challenges of quantization errors and suboptimal ranking. ABRR’s innovative pipeline (comprising temporal distribution smoothing, Wasserstein-based boundary refinement, and quality-driven re-ranking) leverages advanced mathematical tools, including optimal transport theory and Sinkhorn iterations, to achieve precise localization and robust ranking. 
 By operating as a model-agnostic module, ABRR seamlessly integrates with existing TSG models, enhancing their performance without requiring retraining. 
Extensive experiments on Charades-STA, ActivityNet Captions, and QVHighlights demonstrated consistent improvements at stringent IoU thresholds (\textit{e.g.}, IoU=0.7) and consistent gains in mean IoU. 
 Ablation studies confirmed the synergistic contributions of ABRR’s components, while qualitative analyses and computational efficiency evaluations underscored its practical applicability.

 Our work provides a mathematically structured and practically versatile post-processing solution for improving existing TSG systems. Looking ahead, we envision several directions for future research. First, extending ABRR to other video-language tasks, such as video question answering or dense video captioning, could broaden its impact. Second, incorporating temporal context modeling within the optimal transport framework may further improve refinement accuracy for complex, multi-event queries. Finally, exploring adaptive hyperparameter tuning for ABRR’s components could enhance its robustness across diverse datasets and model architectures. ABRR represents a significant step forward in precise temporal localization, paving the way for more accurate and efficient video understanding systems.

\bibliographystyle{IEEEtran}
\bibliography{main}

\end{document}